# Applying Active Diagnosis to Space Systems by On-Board Control Procedures

E.Chanthery, L. Travé-Massuyès, Y. Pencolé, R. De Ferluc, B.Dellandréa *Abstract*—The instrumentation of real systems is often designed for control purposes and control inputs are designed to achieve nominal control objectives. Hence, the available measurements may not be sufficient to isolate faults with certainty and diagnoses are ambiguous. Active diagnosis formulates a planning problem to generate a sequence of actions that, applied to the system, enforce diagnosability and allow to iteratively refine ambiguous diagnoses. This paper analyses the requirements for applying active diagnosis to space systems and proposes ActHyDiag as an effective framework to solve this problem. It presents the results of applying ActHyDiag to a real space case study and of implementing the generated plans in the form of On-Board Control Procedures. The case study is a redundant Spacewire Network where up to 6 instruments, monitored and controlled by the on-board software hosted in the Satellite Management Unit, are transferring science data to a mass memory unit through Spacewire routers. Experiments have been conducted on a real physical benchmark developed by Thales Alenia Space and demonstrate the effectiveness of the plans proposed by ActHyDiag.

## I. Introduction

AUTONOMY is defined as the ability of a system to make its own decisions and act in a changing environment independently of any human intervention. One way to improve autonomy is to endow the system with the ability to evaluate its own health state.

In this paper, the targeted systems are satellites. Autonomous satellites are able to give relevant information to the ground, detect if one of their components is out of order and they can go so far as to decide to reboot some component, switch to a redundant component, or reconfigure themselves. These capabilities require to perform *on-line diagnosis* (fault detection and isolation) and *on-line replanning*. The on-line diagnosis task aims at determining at operating time whether faults have occurred or not within the system and which faults relying on the available set of observations/measurements[1]. On-line diagnosis methods are usually reactive tasks that take as input available measurements provided by the sensors of a physical system and return an estimation of the system's state and its health status. However, a diagnosis method has originally been defined as a process that couples the estimation of the state and the proposal of new specific tests, in particular measurements, that can provide additional information [19]. Diagnosis is then iteratively refined until returning a non ambiguous state estimation. This is a standard procedure in post-mortem diagnosis, which is often formulated as a test sequencing problem [15] but very few works combine diagnosis and testing tasks in the field of on-line diagnosis. The main reason for this is that sensors are often designed for control purposes and control inputs are designed to achieve nominal control objectives. In this setting, available measurements are not designed for health's monitoring and thus may not be sufficient to determine the faults with certainty, resulting in ambiguous diagnoses. The performance of the diagnosis process can however be improved by acting on the system: this is the *active diagnosis* problem. Active diagnosis problems are automated planning problems that aim at designing an admissible sequence of actions (also called a *plan*) whose goal is to refine the diagnosis by reducing the ambiguity without radically changing or degrading the initial mission plan.

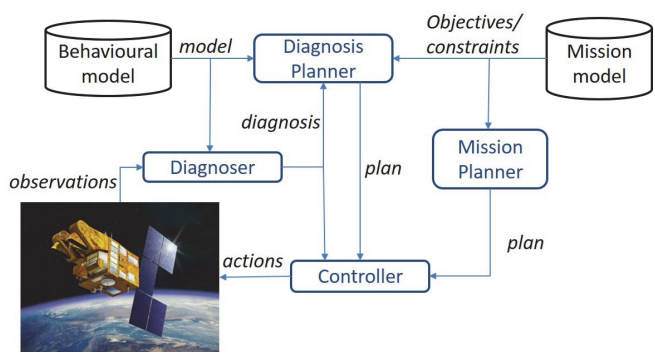

Fig. 1. The active diagnosis scheme.

The needs of an efficient active diagnosis function appear:
- before the occurrence of a fault to know exactly the health status of the satellite,
- when a fault occurs, if the reconfiguration delay is sufficiently long to launch an active diagnosis session, for example when a fault occurs on the payload,
- after the fault, for investigation purposes to get the exact source of the problem.

As illustrated in Fig.1, the active diagnosis scheme interlinks diagnosis and planning. It uses an admissible sequence of actions (or plan**)** to refine the diagnosis without degrading the initial mission plan**.** The space domain implies a set of very specific constraints that limit the applicability of some approaches. In particular, one has to minimize the impact of active diagnosis on the satellite operation. To do this, the active diagnosis algorithm has to guarantee the availability of resources, to maintain the satellite in its current mode, and to not disturb the satellite during a given reconfiguration.

---

[1] These two terms will be used indifferently in the rest of the paper

Moreover, as other permanent on-board diagnosis functions already exist, the active diagnosis algorithm has to get along with other diagnosis mechanisms. Finally, the computational time required by active diagnosis must be consistent with the time constraints inherent with autonomous operations in spatial systems and resource limitations have to be taken into account.

In previous works [5], [3], [7] a formal definition of the active diagnosis problem for discrete-event systems (DES) and an on-board architecture that integrates a diagnosis planner in charge of solving the active diagnosis problem have been proposed. An algorithm was sketched, leaving open the integration of active diagnosis into an on-board architecture including the mission planning process. In [5], [7], this algorithm relies on building a so-called active diagnoser that results from analyzing fault discriminability and precompiling the underlying DES model in order to speed up the on-line search for an active diagnosis plan.

Satellites are complex systems which account for several operation modes and hence involve sensors and actuators showing continuous and discrete event dynamics so they can be typically modelled as hybrid dynamic systems. Hybrid dynamic systems indeed have dynamics that combine continuous-time evolution (i.e. flow) and discrete-time evolution (i.e. switches or jumps) whose description requires merging two mathematical fields, namely differential/difference equations and event-based models.

To solve the diagnosis problem on hybrid systems, the authors of [2] and [4] propose to use the hybrid automata formalism that couples differential equations/difference equations with finite state machines and sign up an original approach based on an abstracted hybrid model in which continuous dynamics are present through specific events. The proposed approach is based on event abstraction and interlinks a standard diagnosis method for continuous systems, namely the parity space method, and a standard diagnosis method for discrete-event systems (DES), namely the diagnoser method [12]. The parity space method provides the means to generate residuals and signatures for each operation mode and these are abstracted as specific events, called signature-events. These events enrich the discrete event model of the system. The diagnoser method is applied to the resulting abstract hybrid system to build a diagnoser able to follow on-line the behavior of the system according to the observable events, including signature-events whose observability signs up mode discernibility [2, 3]. The approach was successfully tested on the ADAPT test bench of NASA Ames with the files of the DX competition 2011 [13]. The active diagnosis approach proposed in [3] relies on this approach.

This paper presents a thorough analysis of the needs and constraints dictated by the space domain for active diagnosis and proposes an effective framework to solve the active diagnosis problem: this framework, in line with the ideas of [3, 5], and [7], is called ActHyDiag. We prove the applicability of ActHyDiag to automatically generate diagnosis plans for a satellite. The experiment starts with fault detection and implements the autonomous application of the generated plan translated into On-Board Control Procedures (OBCP).

The paper is organized as follows. Section II presents some related works on active diagnosis for satellites and autonomous systems and motivates the proposed active diagnosis approach, clarifying the contributions of the paper. Section III presents the Spacewire case study. Section IV presents ActHyDiag, the active diagnosis framework. Section V applies this framework to the case study. Finally, Section VI concludes the paper by discussing the results and outlining perspectives for future work.

## II. RELATED WORK

### A. Active Diagnosis Frameworks for DES or hybrid systems

In this paper, actions are of discrete type and can be modelled as discrete events but they may require to synthesize continuous control inputs when indicating a mode transition in a hybrid system. Nevertheless, this latter problem is out of the scope of this paper.

The method presented in this paper is based on proposing actions to enforce diagnosability like [3], [5], and [7]. State space regions of poor diagnosability are accepted because most real systems are designed this way. When the system is in an ambiguous diagnosis state, active diagnosis enforces actions to drive the system towards regions with improved diagnosability, achieving the highest possible diagnosis refinement. This approach adopts the perspective that the operation of the system can be suspended to run an active diagnosis session. The active diagnosis actions hence do not interact with actions dedicated to control.

With the same idea, the method proposed in [20] is based on a set of models (one per fault) that are used to predict the future output of the system in each situation. At time $t$, the active diagnosis problem is formulated as a mixed integer optimization problem using the Mixed Logical Dynamical framework. It provides an input sequence that aims at distinguishing the behavior of each model. The input is applied to the system and the whole procedure is repeated at time $t+1$ until the estimated output of the different systems are all different. The decision function compares the output and the estimated output in order to decide about the fault candidate. This approach is not easily applicable to space systems because the generated plan interlinks continuous control and discrete control actions. It seems more appropriate to separate these two control levels as proposed in this paper. The complexity of the method also severely limits its applicability on-board.

Three other main approaches for active diagnosis can be found in the literature. The first is based on *preventing actions to forbid non diagnosable regions* [17]; the second claims that a *combined approach* that prevents and proposes actions is necessary [14]; the third takes benefit of flexible mission plans to select one that is useful for diagnosis purposes. It is also known as the *pervasive diagnosis* approach [11].

Following the first approach, the authors of [17] use control actions to alter the diagnosability properties of a given discrete event system. Active diagnosis is formulated as a supervisory control problem [16] and the controller is designed so that specific actions that may drive the system into non diagnosable regions are forbidden. The system is hence maintained "actively" diagnosable and its diagnoser produces non ambiguous outputs. In other words, active diagnosis is achieved by preventing inappropriate control actions. The main drawback of this method is that preventing actions is not suitable at all to the space domain. Indeed it is important to maintain all actions applicable for the purposes of the mission.

A combined approach is proposed in [14] as a solution where active diagnosis is enabled to avoid ambiguities by selectively blocking or executing actions. It is formulated in an event-based framework where event-based diagnosers are synthesized and they can determine if the system is diagnosable through passive or active diagnosis. This method obviously inherits from the drawback of the previous method as preventing actions may be a problem.

Pervasive diagnosis has been proposed in [11] to construct informative mission plans that simultaneously achieve mission goals and provide diagnostic information. A heuristic search algorithm has been proposed for generating these informative mission plans. In other words, pervasive diagnosis optimally achieves as much diagnosis as possible during the mission, adding a diagnosis capability without disrupting the mission. Active diagnosis and mission can, therefore, take place simultaneously, leading to interlinked diagnosis and mission planning. This method gives the mission plan a huge impact on the diagnosis. This strategy is applicable only when the cost of failure is low compared with the cost of stopping the mission to perform diagnosis, which is hardly the case for space systems. Indeed, failure may result in losing the satellite and canceling the mission anyway. Pervasive diagnosis is hence interesting in application domains in which failures do not lead to catastrophic and non-reversible situations, which is not the case in the space domain.

*B. Motivations in relation with current FDIR in Space Systems*

The main health management constraint for satellites is to avoid the system's loss at any cost. The on-board Fault Detection, Isolation and Recovery (FDIR) strategy aims at detecting faults, isolating them and reconfiguring the satellite either in the previous operational mode (fail-op), or in one of the safe modes (fail-safe) along the following strategy:

- Level 0 fault: the fault is detected by a device and is automatically reconfigured; the system returns to its previous operational mode.
- Level 1 fault: the fault is detected by a device and is reconfigured by the Data Handling Subsystem (DHS) (generally the On-Board Computer or OBC), usually by using hardware redundancy; the system returns to is previous operational mode.
- Level 2 fault: the fault is detected by the DHS thanks to inconsistency of Attitude and Orbit Control System (AOCS) sensors. These faults relate to late detections of anomalies without the ability to distinguish a clear origin. The passivation of these faults relies on putting the satellite in one of its safe modes.
- Level 3 fault: a fault in the onboard software generates a watchdog alarm. This type of faults is managed by rebooting the on-board software with or without an OBC change. The passivation of these faults relies on putting the satellite in one of its safe modes.
- Level 4 fault: these faults trigger external major alarms like a too weak battery level, a presence of light in unauthorized places, too high angular velocity… The passivation of these faults relies on putting the satellite in one of its safe modes.

According to space experts feedback, at level 0 and 1 faults are by definition detectable at the equipment level and diagnosis is quite obvious. The associated FDIR strategy is implemented either within the equipment itself or within the on-board software. Higher-level faults are more problematic because their origin is usually unknown and may require improvised, ground-based procedures where operators typically conduct the tests by switching off each equipment one by one. These are exactly the type for faults for which the proposed active diagnosis method is useful: instead of switching off each equipment one by one on ground-request, the idea is to use an OBCP instead. Let us notice that the current FDIR process is highly deterministic depending on the analysis of unit failure, on the definition of monitoring parameters, isolation and reconfiguration processes. Some complex faults (of level 2 to 4) can only be monitored at system level, which prevents the actual identification of the failed unit, and triggers a global reconfiguration, often resulting in loss of availability of the overall spacecraft from several hours to several days. The proposed innovation enables to increase the level of investigation to add new means of failure identification and therefore to optimize system availability.

The main contributions of this paper are the following. First, an active diagnosis method is proposed that maintains all actions applicable for the purpose of the mission and that does not interfere with any control strategy of the system. Second, the active diagnosis solution is designed so as to manage non trivial types of faults, integrating new criteria, such as cost of actions. Third, the active diagnosis function is integrated within the existing FDIR strategy via the OBCP mechanism. In the case study, the existing FDIR OBCP is implemented as a state machine managing both fault detection and reconfiguration of the Spacewire Network. The proposed active diagnosis solution can be easily translated into the same OBCP format, allowing for an easy comparison. Moreover, real time constraints of the space system are respected: for example the Spacewire's dynamics impose a frequency of 1Hz for the acquisition of each equipment health status.

III. CASE STUDY

*A. Case Study Choice*

The three main case studies for the active diagnosis experimentation that were identified are: instrument monitoring through a Spacewire Network, thruster fault diagnosis with 16 or 24 thrusters, and solar cell calibration methods. Handling thruster fault diagnosis and solar cell calibration implies dealing with continuous and discrete dynamics. These systems hence require a hybrid model coupling differential equations/difference equations and finite state machines. We decided that this makes complex the experimentation without much added value and we opted for the instrument monitoring through a Spacewire Network. On the other hand, the instrument monitoring through a Spacewire Network presents the following advantages: i) it offers cases where the faulty component cannot be easily identified, ii) it remains simple which allows to validate the outputs of the process by hand, and iii) it can be modelled as a discrete event system, which can also be dealt by ActHyDiag by ignoring continuous dynamics. Last but not least, the dynamics and active diagnosis strategies are compatible with the OBCP mechanisms.

This case-study is further described in the next sections

*B. Spacewire Case Study Presentation*

This case-study is based on a redundant Spacewire Network topology where up to 6 instruments, monitored and controlled by the on-board software hosted in the Satellite Management Unit (SMU), are transferring science data to a mass memory unit called the Payload Data Handling and Transmission (PDHT) through Spacewire routers. The data transferred in the PDHT is downloaded later to the ground when possible. This architecture is illustrated in Fig. 2.

The dataflow within the network therefore includes science data from instruments to PDHT, telecommands from SMU for instrument control, and telemetries from instruments to SMU for monitoring and control. An important aspect is the configuration of the routers, that binds inputs and outputs communication ports for the communications with a specific protocol: a router output port is allocated to a transaction until this transaction ends, and cannot be used until it is released. Spacewire Network relies in flow control token exchanges guaranteeing that until the receiver does not read data from its reception port, no data can be sent that would overflow the communication buffers. Likewise, the routers are relying on wormhole (blocking) mechanism and congestions can happen. A massive congestion can be especially tricky to solve since several causes can generate it and there is no easy mechanism to isolate its origin.

The case study focuses on the nominal part of the network (on the left of Fig. 2), and the interest is put on a fault that can occur on any instrument, and propagate across the network leading to network saturation. Instruments can be ON or OFF. In ON mode, the nominal behavior of an instrument is to send science data to the mass memory through a reasonable traffic, to execute some telecommands coming from the SMU, and to send to the SMU some housekeeping telemetries.

The faulty behavior of an instrument is a "babbling idiot" behavior. The instrument produces very big science packets that will saturate the network: the mass memory will not be able to store all incoming science data. When this situation occurs, SMU is still able to send commands to instruments (request of a healthy status, mode change request, …) but the instruments may not respond if their emission buffers are full.

The SMU controlling the network can configure the Spacewire router ports one by one. One possible action in case of network saturation is therefore to switch OFF and then ON a router port associated with a specific data transaction, leading to the discard of the packet.

Health state of the overall system is monitored by the SMU and obtained by instrument health status observed via a periodic (1Hz) health check request and response between the SMU and each instrument. When the network is saturating due to a faulty instrument, some or all instruments may not answer to the health check request.

When the network is saturating, diagnosis intuitively consists in progressively switching OFF the router ports associated to the instruments until the network retrieves a healthy status, and health status of the remaining instruments are able to reach the SMU node. This active diagnosis plan will be generated automatically by ActHyDiag and implemented as an OBCP running on the SMU

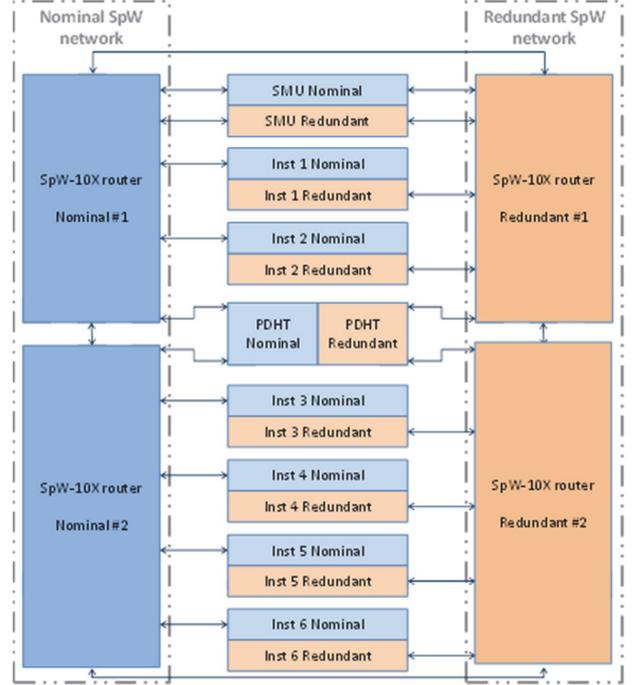

Fig. 2. Spacewire Network architecture

## IV. ACTHYDIAG: AN ACTIVE DIAGNOSIS FRAMEWORK

*A. Modeling Formalism*

ActHyDiag is an extension of the software HyDiag that implements diagnosis for hybrid systems [4, 2, 8]. Hybrid systems are modelled as hybrid automata [10]. Formally, a hybrid automaton is defined as a tuple $S = (\zeta, Q, \Sigma, T, C, (q^0, \zeta^0))$ where:

- $\zeta$ is a finite set of continuous variables that comprises input variables $u(t) \in \mathbb{R}^{n_u}$, state variables $x(t) \in \mathbb{R}^{n_x}$, and output variables $y(t) \in \mathbb{R}^{n_y}$.
- $Q$ is a finite set of discrete system states.
- $\Sigma$ is a finite set of events.

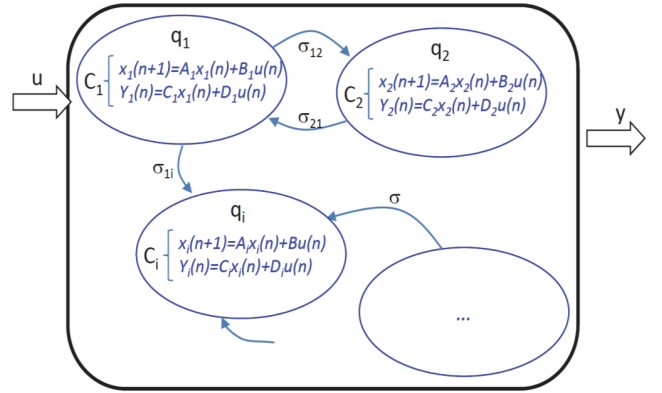

Fig. 3. Example of a hybrid system.

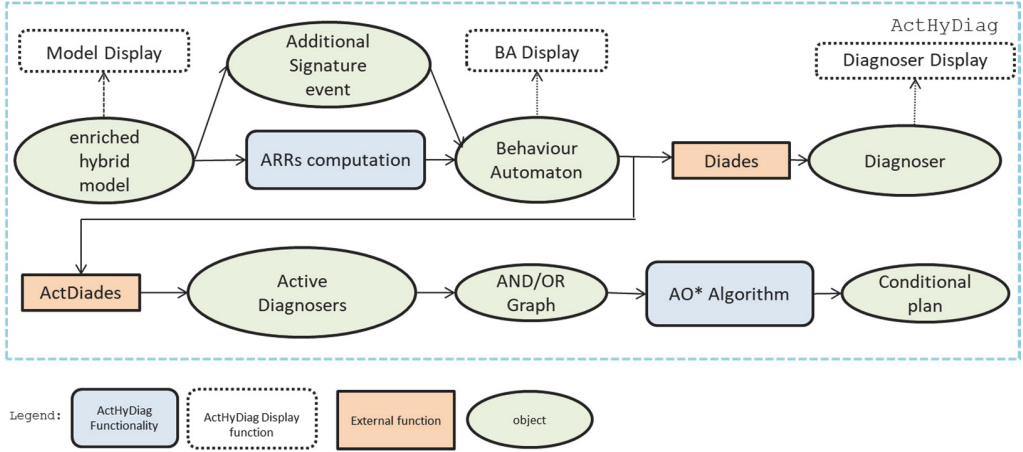

Fig. 4 ActHyDiag architecture

- $T \subseteq Q \times \Sigma \rightarrow Q$ is the partial transition function between states.
- $C = \bigcup_{q \in Q} C_q$ is the set of system constraints linking continuous variables.
- $(\zeta^0, q^0) \in \zeta \times Q$, is the initial condition.

Each state $q \in Q$ represents a behavioral mode characterized by a set of constraints $C_q$ that model the linear continuous dynamics (defined by their representations in the state space as a set of differential and algebraic equations). A behavioral mode can be nominal or faulty (anticipated faults). The unknown mode can be added to model all the non-anticipated faulty modes. The discrete part of the hybrid automaton is given by $M = (Q, \Sigma, T, q^0)$, which is called the underlying DES. $\Sigma$ is the set of events that correspond to discrete control inputs, autonomous mode changes and fault occurrences. The occurrence of an anticipated fault is modelled by a discrete event $f_i \in \Sigma_f \subseteq \Sigma_{uo}$, where $\Sigma_{uo} \subseteq \Sigma$ is the set of unobservable events. $\Sigma_o \subseteq \Sigma$ is the set of observable events. Transitions of $T$ model the instantaneous changes of behavioral modes. The continuous behavior of the hybrid system is modelled by the so called underlying multimode system $\Xi = (\zeta, Q, C, \zeta^0)$. The set of directly measured variables is denoted by $\zeta_{OBS} \subseteq \zeta$. An example of hybrid system modelled by a hybrid automaton is shown in Fig. 3. Each mode $q_i$ is characterized by state matrices $A_i, B_i, C_i$, and $D_i$.

### B. Architecture and Principles

ActHyDiag operates via the following steps represented in Fig. 4: the first step is the input model that requires to fill out the fields previously explained (modes, events). A file that indicates the events of $S$ that are actions, as well as their respective cost is added. This is called the *enriched hybrid model*. ActHyDiag then generates Analytical Redundancy Relations (ARR) for each mode using the well-known parity-space approach [9] to build the *Behavior Automaton* (BA). The idea is to capture both the continuous dynamics and the discrete dynamics within the same mathematical object BA. The enriched model is completed with specific observable events, called *signature-events* that are generated from the mode signatures provided by the corresponding ARRs, or directly by an expert. A specific transition labelled with a specific signature-event is introduced between two modes when they have different signatures. The resulting automaton is $BA(S) = (Q_{beh}, \Sigma^{BA}, T_{beh}, q_{BA}^0)$, where $Q_{beh}$ is the union of the finite set of discrete system states and the states that model the continuous reaction after the occurrence of a discrete event, $\Sigma^{BA}$ is the alphabet that contains discrete events and events modelling the signature switches, $T_{beh}$ the partial transition function and $q_{BA}^0$ the initial state.

Diagnosis is performed thanks to a specific finite machine called a *diagnoser*. The diagnoser is built from the BA following the approach described in [12]. The task of building such diagnoser is not easy because it requires to browse the entire graph representing the BA automaton. To this end, the tool DiaDES[2] is used and allows us to generate the diagnoser automatically. The user can run in parallel a simulation of the system and he/she can display the following data: inputs, outputs, events, belief state of the diagnosis, etc.

Active diagnosis is performed thanks to an *active diagnoser*. Based on the BA, ActHyDiag computes an active diagnoser that is able to predict whether or not a fault can be diagnosed with certainty by applying an action plan from a given ambiguous situation [7]. From this active diagnoser, a planning domain in the form of an AND/OR graph can be extracted.

At runtime, the diagnosis might be ambiguous. An active diagnosis session can be launched as soon as the active diagnoser can assess that the current faulty situation is discriminable by applying some actions. If the active diagnosis session is launched, an AO∗ algorithm starts and computes a conditional plan from the AND-OR graph that optimizes an action cost criterion. It is important to notice that the active diagnosis plan issued by ActHyDiag only displays discrete actions. In particular, it is assumed that if it is necessary to guide the hybrid system towards a state based on continuous control, the synthesis of control laws must be performed independently. However, this feature is not required for the selected case study.

---

[2] DIADES is a software from LAAS-CNRS, Toulouse, France. Documentation and download available on http://homepages.laas.fr/ypencole/DiaDes/.

*C. Active Diagnoser*

This section describes the construction of the so-called active diagnoser. The early principles of the active diagnoser were introduced in [3][5]. The aim of this data structure is to precompile the knowledge that is necessary for the generation of the active diagnosis plans from the behavioral automaton (BA) described in the previous section [6]. Here we recall the definition of the active diagnose and provide the details that are necessary to understand how to effectively compute it.

The active diagnoser is an automaton whose transitions are labeled with an observable event $o \in \Sigma_o^{BA}$ from the BA. The active diagnoser is deterministic and complete, i.e. each state contains exactly $|\Sigma_o^{BA}|$ output transitions and any event of $\Sigma_o^{BA}$ is the label of one of them. Formally speaking, the active diagnoser $\Delta$ is the automaton $\Delta = (Q_\Delta, \Sigma_o^{BA}, \delta, q_\Delta^0, \tau)$ where:
- $Q_\Delta$ is a finite set of states;
- $\Sigma_o^{BA}$ is the diagnoser alphabet, i.e. the observable events from the BA;
- $\delta: Q_\Delta \times \Sigma_o \rightarrow Q_\Delta$ is the transition function;
- $q_\Delta^0 \in Q_\Delta$ is the initial state;
- $\tau: Q_\Delta \times \Sigma_f \rightarrow tags(\Sigma_f)$ is the tag function.

The set of states $Q_\Delta$ and the transition function $\delta$ are defined as follows. Let $paths(Q_1, \sigma, Q_2)$ denote the set of transition paths of the BA such that the source is in $Q_1$, the target is in $Q_2$, the observable part is exactly the sequence $\sigma$ and the last transition is observable, which therefore means that so is the last event of $\sigma$. Each state $q$ of the active diagnoser is mapped with a couple $(BA(q), Tags(q))$:
1. the set $BA(q)$ is a subset of states of BA;
2. the vector $Tags(q) \in \prod_{i=1}^n tags(f_i)$ with $tags(f_i)$, $f_i \in \Sigma_f = \{f_1, \ldots, f_n\}$, being the set of following tags:
$\{f_i - safe, f_i - sure, f_i - discriminable, f_i - nondiscriminable, f_i - nonadmissible\}$.

The set of states $Q_\Delta$ and the transition function $\delta$ are then defined by induction as follows.
1. Let $Q_\Delta^0 = \{q_\Delta^0\}$ where $q_\Delta^0$ is associated with $BA(q_\Delta^0) = \{q_{BA}^0\}$ where $q_{BA}^0$ denotes the initial state of the BA and $Tags(q_\Delta^0) = (f_1 - safe, \ldots, f_n - safe)$.
2. Let $\delta^i: Q_\Delta^i \times \Sigma_o \rightarrow Q_\Delta^{\delta i}$ for any $i \geq 0$ be such that, if $BA(q) = \emptyset$ then $\delta^i(q, o) = q$ for any event $o$ and $Tags(q) = (f_1 - nonadmissible, \ldots, f_n - nonadmissible)$.
Otherwise $\delta^i(q, o) = q'$ and $BA(q')$, $q' \neq q$, is defined as the set of BA states such that $paths(BA(q), o, BA(q'))$ is maximal. Note that $BA(q')$ might be empty (in that case $paths(BA(q), o, BA(q'))$ is also empty), due to the completeness of the diagnoser. For any fault event $f_i$, let $Tags(q, i)$ denote the i-th element of the vector $Tags(q)$.

    a. $Tags(q', i) = f_i - safe$ if for any observable sequence $\sigma$ such that $paths(BA(q_\Delta^0), \sigma, BA(q'))$ is not empty then none of these paths contains the event $f_i$.
    b. $Tags(q', i) = f_i - sure$ if all of the paths in $paths(BA(q_\Delta^0), \sigma, BA(q'))$ contain the event $f_i$.
    c. If only part of the paths in $paths(BA(q_\Delta^0), \sigma, BA(q'))$ contain the event $f_i$ then $Tags(q', i) = f_i - nondiscriminable$ if there is no observable sequence $\sigma\sigma'$ such that:
        i. there exists a subset $Q'$ of the BA's states that maximizes the set $paths(BA(q_\Delta^0), \sigma\sigma', Q')$, and
        ii. the set $paths(BA(q_\Delta^0), \sigma\sigma', Q')$ is not empty, and
        iii. either any of these paths contain the event $f_i$ or none of them contains it.

Finally, $Tags(q', i) = f_i - discriminable$ if $Tags(q', i)$ is not $f_i - nondiscriminable$.

3. Let $Q_\Delta^{i+1} = Q_\Delta^i \cup Q_\Delta^{\delta i}$.

If $\delta^i$, $i \geq 0$, is recursively applied from $Q_\Delta^0$, as the BA is finite, there exists a step $n$ such that $Q_\Delta^{n+1} = Q_\Delta^n$. The set of active diagnoser states is finally defined as $Q_\Delta = Q_\Delta^n$ and the transition function is $\delta = \delta^n$. Finally, for any state $q$ and any fault $f_i$, $\tau(q, f_i) = Tags(q, i)$.

To sum up, the precompilation of the BA into the active diagnoser computes a tag associated to each diagnoser state for each fault $f_i$. The tag determines whether an active diagnosis plan exists to disambiguate between the presence and the absence of each fault $f_i$. This is summed up in Table 1.

*D. Generation of diagnosis plans*

For finding the optimal plan, a best-first search algorithm that explores a graph built on-the-fly from the active diagnoser is used. This graph is an AND-OR tree (Nilsson,1998), where OR nodes correspond to a sequence of observations and AND nodes correspond to a sequence of actions. The algorithm tries to avoid as much as possible building the entire AND-OR tree because it may require too much memory and computation time resources.

In the main algorithm (AO) sketched in Fig. 5, the RootNode is an ambiguous state where the active diagnosis starts. It has to be tagged as $F - discriminable$ at least for one fault for being considered as solvable. The active diagnoser is enriched by costs associated to each action. Observations are supposed to have 0 cost. Several criteria and exploration options are defined. It is possible to explore all possible AND nodes or to explore only one branch of the tree (by depth-first search). It is also possible to explore only the "cheapest" AND nodes. These solutions are useful when the solution space is very large.

The AO_Algorithm used to find an optimal conditional plan for refining the diagnosis is detailed in Fig. 6 [15]. As said before, functions CreateANDSuccessors and CreateORSuccesors create the AND-OR graph on-the-fly knowing the CurrentNode and the active diagnoser. Infinite or too expensive branches are pruned during the search.

| Tag | Meaning | Interest for active diagnosis |
|---|---|---|
| $f_i$- nondiscriminable | It is impossible to determine if fault $f_i$ has occurred or not, even if we act and wait for more observations. | NO: the active diagnosis problem has no solution for $f_i$. |
| $f_i$- discriminable | From the current state, it exists at least some potential future observable sequences that can decide with certainty whether $f_i$ has occurred or not. | YES : from the current state, there might exist an action plan that will produce one of the expected future observable sequences that can assert the presence or the absence of the fault $f_i$. |
| $f_i$- safe | $f_i$ has not occurred. | NO: the diagnosis is not ambiguous |
| $f_i$- sure | $f_i$ has occurred. | NO: the diagnosis is not ambiguous |

Table 1. Tags, meanings and interest for active diagnosis

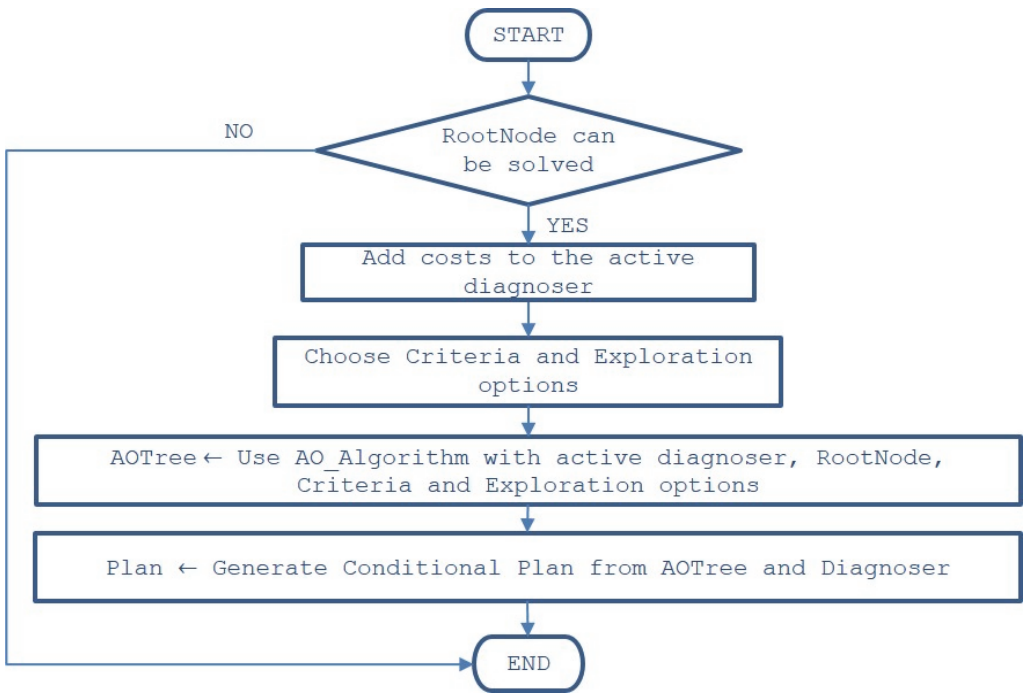

Fig. 5: AO main Algorithm flowchart

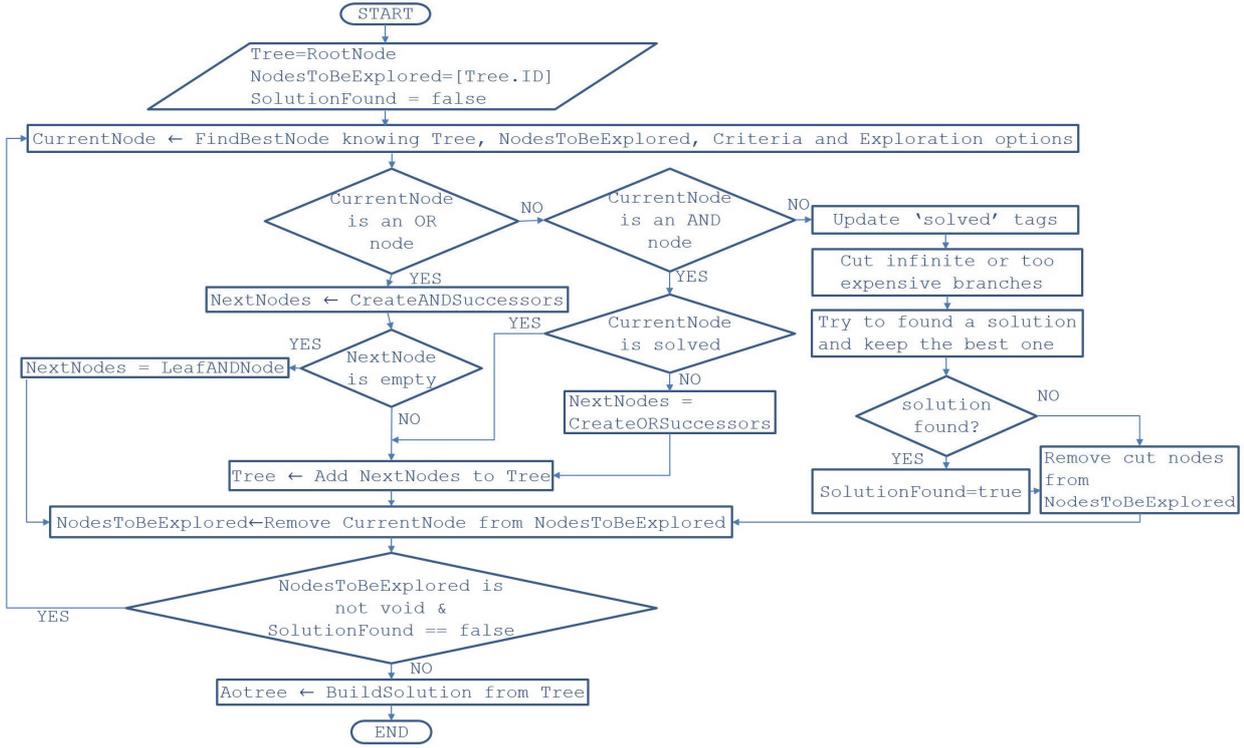

Fig. 6: AO_Algorithm

## V. APPLICATION TO THE SPACEWIRE

### A. Model of the SpaceWire

SpaceWire components are modeled as automata and implemented in the Supremica software [1]. Supremica is an integrated environment for verification, synthesis and simulation of discrete event systems. We only use it to synchronize the SpaceWire component models.

The models of the 3 instruments and the associated router ports are illustrated in Fig. 7 Switch on and switch off actions were removed from the model. Indeed, for active diagnosis, the only actions required are closing and opening the flows associated to instruments. The global model is obtained by synchronization. One difficulty is the number of transitions resulting from synchronization.

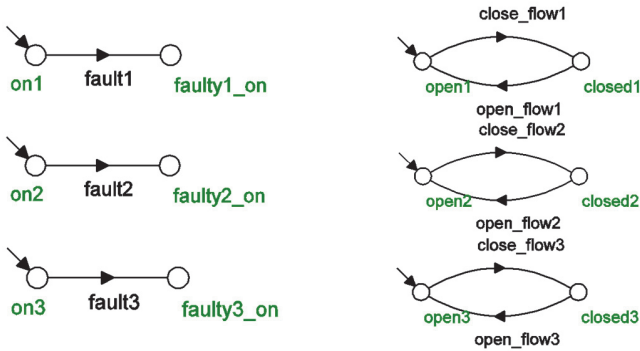

Fig. 7: Supremica automata for instruments and SpaceWire router

The model of the full case study excluding the actions corresponding to RMAP requests for verifying the health status ("check" events) includes 64 states, 9 events and 288 transitions. Once the complete model is generated with Supremica, it is necessary to import it in ActHyDiag. A Matlab function has been developed to import the .xml file generated by Supremica into a .mat file understandable by ActHyDiag. Then the cost of each action must be added before launching the active diagnosis plan search algorithm. The "check" operation (2 transitions and one state) is added to the model for each relevant state. The strategy for active diagnosis is to consider the overall state of the network as saturated or unsaturated. The overall status is determined from responses to the health status returned by the instruments whose ports are open.

Our experiment aims not only to provide solutions for the SpaceWire benchmark, but also to study the feasibility of applying active diagnosis in cases where the number of components is increased. In this context, our goal has been to develop a method for generic modelling. This method can be summarized with the following steps: 1) model components in Supremica, 2) synchronize component automata, 3) export .xml file then create .mat file and import it in ActHyDiag, 4) add costs, 5) generate active diagnosis plans from relevant nodes in the diagnose.

### B. Active Diagnoser

The models were imported into Matlab and the diagnosers were generated up to 6 instruments.

The particularity of the active diagnoser for the SpaceWire is that nodes are tagged either *F-discriminable* or *F-sure*, i.e. nodes tagged *F-nondiscriminable* or *F-safe* do not exist. For the example of 3 instruments, the active diagnoser has 304 nodes; for 4 instruments it has 5344 nodes. The objective is to find the most relevant node for an active diagnosis session. A priori, it is the node where all the flows are open and in which the network is detected saturated. For the active diagnosis tests, we launched the active diagnosis algorithm when the diagnoser was in that node.

*C. Translation into OBCP*

In order to validate experimentally the conditional plans obtained by the proposed active diagnosis method, it is necessary to derive corresponding OBCP procedures. Two aspects are considered during this step:

- Adding system dynamic aspects: they are taken into account by adding some delays as time was not modelled in the plan generation,
- Translation into the target language: the conditional plan is defined as pseudo code and needs to be translated into Java, using the available services.

For the Spacewire case-study, the entire conditional plan is contained in a Java function called `diagnose_network()` that inherits from the plan structure, with each action or observation translated as a call to an external service. Examples of translation of conditional plan lines and java service calls are given in Table 2.

| Conditional Plan | Java service call |
|---|---|
| close_flow1 | monitor_port_state(1, false); |
| open_flow3 | monitor_port_state(3, true); |
| Check | monitor_scan(); |
| if check OK | if (monitor_failing_nb == 0) |
| if check KO | if (monitor_failing_nb != 0) |
| fault1 sure and fault2 sure and fault3 sure END | diag_result = 111; return diag_result; |
| [missing timing info] | monitor_tempo(); |

Table 2. Conditional plan into java service call translation

The global function `diagnose_network()` returns an integer value containing the result of the diagnosis as presented in Table 3.

| Diag_result value | Result |
|---|---|
| 111 | fault1 sure and fault2 sure and fault3 sure |
| 11 | fault1 unsolvable and fault2 sure and fault3 sure |
| 101 | fault1 sure and fault2 unsolvable and fault3 sure |
| 110 | fault1 sure and fault2 sure and fault3 unsolvable |
| 100 | fault1 sure and fault2 unsolvable and fault3 unsolvable |
| 1 | fault1 unsolvable and fault2 unsolvable and fault3 sure |
| 10 | fault1 unsolvable and fault2 sure and fault3 unsolvable |

Table 3. Result of diagnosis

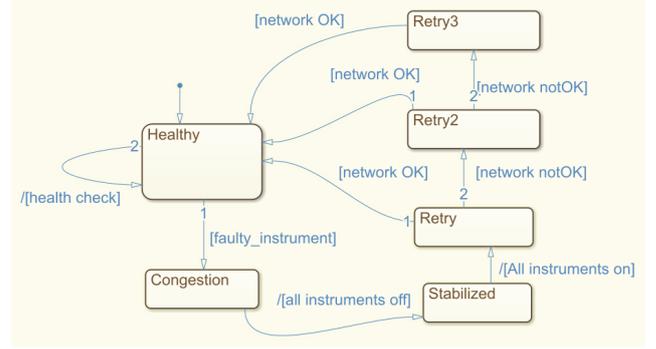

Fig. 8: Original statechart for the Spacewire Network FDIR

The objective was then to inject this java active diagnoser function within the existing prototype of the case-study that already implements an FDIR strategy based on OBCP. The original OBCP was in the form of a state machine implementing both fault detection and reconfiguration of the Spacewire Network, as illustrated in Fig. 8.

*D. Scenarii and Results*

The system starts in a healthy state where the network is not yet saturated. The FDIR OBCP performs periodic monitoring of the network by sending health check requests to the instruments that have to answer subsequently. When a faulty instrument produces a huge packet, all the instruments data flows towards the mass memory are suspended and the entire network gets saturated, preventing health check responses to be sent by the instruments. As soon as several responses are missing, the OBCP goes to the "congestion" state. In this state, the router ports corresponding to the instruments that did not answer the health check request are closed. In fact all ports need to be closed as the full network is saturating very quickly due to the high data rates. After full closure of the ports, the OBCP enters in the "stabilized" mode where router ports are re-open one by one. As the simulated fault is transient, there is no more faulty instrument in this phase. However, because of a bug in the hardware used for the demonstration (Spacewire router USB connexion misbehavior), the confirmation that the network is OK may require up to three retries.

In order to validate the active diagnosis elaborated as a conditional plan, the OBCP is modified as follows. When the congestion of the network is detected, the active diagnosis function is called. When the active diagnosis function has identified the faulty instrument, it relies on the original FDIR OBCP processes to proceed with the network reconfiguration (the active diagnosis does not include reconfiguration aspects).

The new OBCP can easily be loaded into the prototype, and several fault injection scenarios can be run. Two of them are reported below.

In Scenario 1, a transient fault on instrument 3 was tested. The active diagnosis OBCP was tested, but the diagnosis result was not correct: instrument 3 was the faulty one, but the diagnose_network function indicated that instrument 1 was faulty. A deep analysis of the packets exchanged showed that the mismatch was related to the limitation of the prototype and lack of observables on the Spacewire Network. This constitutes a perfect illustration of the difficulty to find root causes of Spacewire Network saturation and also an illustration of the difficulty to model reality.

In Scenario 2, a persistent fault on instrument 3 was tested. This fault was simulated physically, preventing the faulty instrument to provide a health check response after emission of the big packet. In this case, the active diagnosis OBCP identified correctly the instrument 3 as the faulty one.

In its current state, the testbed does not allow us to experiment multiple faults in the instrument set. It has to be improved for testing multiple faults. However, the algorithm considers that multiple faults are possible: active diagnosis plans are then more complex than those intuitively expected by someone assuming simple faults.

Concerning the reconfiguration step, this approach adopts the perspective that the operation of the system is suspended to run the active diagnosis session. The active diagnosis actions hence do not interact with actions dedicated to control. Note also that active diagnosis is designed to isolate ambiguous diagnoses, not to be fault tolerant. The proposed solution enables to recover a failed system after a transient failure. This transient effect might result in data loss. Since the proposed innovation is external to the communication system, it cannot compensate for the data loss. This could only be achieved through an increase in QoS (Quality Of Service) for the overall communication either from the communication system itself (through hot redundancy or retry processes) or at higher level through packet acknowledgement at transport level (like TCP for instance) or higher (applications monitoring the effect of commands, like PUS (Packet Utilisation Standard) services for instance in space applications). A perspective of this work is then a higher use of deterministic communication protocols in networks that allocate good bandwidths with guaranteed quota.

## VI. Conclusion And Future Work

This article highlights the interest of active diagnosis encoded as an OBCP. The tools implemented can handle complex nontrivial cases for system engineers, incorporating interesting dimensions such as cost criteria. Interestingly, the translation of the generated conditional plan as an OBCP is quick and easy (search / replace text). The OBCPs therefore are an ideal format for this type of functionality, and the proposed concepts perfectly meet the needs of active diagnosis.

Three industrial obstacles are highlighted in this study: i) the length of the conditional plan can become very large depending on the number of instruments, making it impossible for the implementation as an OBCP as it would be too expensive in terms of on-board memory. ii) the modelling of the considered system may be tedious. iii) validation of the generated plans can be complex and require a large number of test cases. This later issue can be made easier by analyzing behavioral symmetries of the considered system. In any case, theoretical validation is an issue that should be addressed in the future.

Another issue concerns real-time aspects. For the case study, the execution performance of the OBCP is compatible with the dynamics of the Spacewire Network (health status at 1 Hz). In the testbed, the OBCP is the only procedure run in the node hosting the flight software. There is no conflict between the OBCP and critical functions. Recommended principles and mechanisms have therefore not been experienced through the testbed. This remains a major point which will determine the feasibility of the implementation of the active diagnosis on-board. This issue could ultimately be resolved by system choices. Tests that have been done during the study are not representative of a real embedded target, so the results aspects, like processing power/time or memory required are not relevant. We are currently working on a more representative testbed. It will be used for demonstrating real-time aspects of the implementation. The proposed evolution in on-board FDIR handling contains several steps of system optimization. Through these steps, a progressive implementation is possible depending on the on-board capabilities. During the study it has been demonstrated that a subset of the active diagnostic processes can already be successfully implemented to manage complex communication networks. Further subsystems could also benefit from this innovation in future avionics with enhanced processing capabilities, for example thruster faults or faults in the calibration of the flexible modes of solar panels.

Future work will enable to process and experience the deterministic aspects of active diagnosis by OBCP (performance runtime, characterization of the worst cases, integration into the complete flight software...). The scaling issues will also be addressed: an interesting solution may be to consider the distribution of the generation of the conditional plan in a distributed monitoring framework.

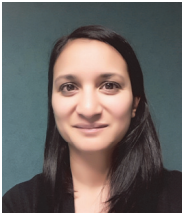

**Elodie Chanthery** received the engineering degree and the master degree from the Ecole Nationale Supérieure d'Electrotechnique, d'Electronique, d'Informatique, d'Hydraulique et des Telecommunications, Toulouse, France, in 2002, and the Ph.D. degree in mission planning for autonomous vehicles from the Ecole Nationale Supérieure de l'Aéronautique et de l'Espace-SupAero, Toulouse, in 2005. She received the Habilitation à Diriger des recherches from the Institut National Polytechnique de Toulouse in 2018. She is an Assistant Professor of Control with the Institut National des Sciences Appliquées, Toulouse. Her research activities are conducted in LAAS-CNRS in the field of complex hybrid systems, focusing on the links between diagnosis and other tasks such as planning or prognosis.

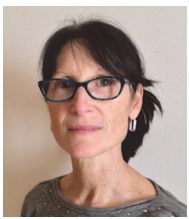

**Louise Travé-Massuyès** holds a position of Directrice de Recherche at Laboratoire d'Analyse et d'Architecture des Systèmes, Centre National de la Recherche Scientifique (LAAS-CNRS), Toulouse, France. She graduated in control engineering from the Institut National des Sciences Appliquées (INSA), Toulouse, France, in 1982 and received the Ph.D. degree from INSA in 1984.

Her main research interests are in dynamic systems diagnosis and supervision with special focus on model-based and qualitative reasoning methods and data mining. She has been particularly active in establishing bridges between the diagnosis communities of Artificial Intelligence and Automatic Control. She has been responsible for several industrial and European projects and published more than 250 papers in scientific journals and international conference proceedings and 4 books. She is coordinator of the Maintenance & Diagnosis Strategic Field within the French Aerospace Valley World Competitiveness Cluster, and serves as the contact evaluator for the French Research Funding Agency. She serves as Associate Editorial for the well-known Artificial Intelligence Journal. She is member of the International Federation of Automatic Control IFAC Safeprocess Technical.

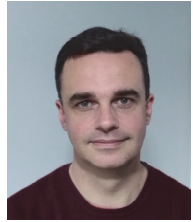

**Yannick Pencolé** is a CNRS researcher at LAAS-CNRS since 2006 and is the leader of the DISCO research team (Diagnosis and Supervisory Control). From 2003-2006, he was a research fellow at the Australian National University, Canberra, Australia in the DPO group (Diagnosis, Planning and Optimization). He received a PhD degree in 2002 on Model-based Diagnosis from the University of Rennes. His research interests include model-based diagnosis of large scale discrete-event systems and logical systems, meta-diagnosis reasoning, active diagnosis, diagnosability, and system maintenance.

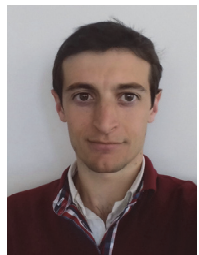

**Régis de Ferluc** is an Innovation R&D project manager at Thales Alenia Space, Cannes, France. He graduated in Real-Time and Embedded Systems from the Institut National des Sciences Appliquées (INSA), Toulouse, France, in 2010. He started his carrer in Thales Alenia Space as a member of the on-board software R&D team, and participated to numerous ESA or CNES R&D studies, and to several European Projects, dealing with satellite on-board software and satellite monitoring and control. In 2016, he took the responsibility to coordinate software R&D activities in Thales Alenia Space, Cannes.

His main research interests are Real-time Operating systems, Time and Space partitioning, Modelling tools and Processes to suport on-board software design, avionics design, and FDIR (Fault Detection Isolation and Recovery). In this latter area, he published several papers in scientific or industrial conferences proceedings.

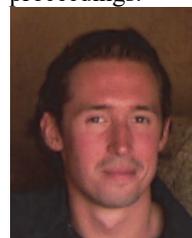

**Brice Dellandrea** is currently head of the R&T and advanced projects department for satellite platforms within Thales Alenia Space Cannes. He leads new developments and technologies derisking in the fields of data handling, communications, power and guidance, navigation & control.

He is R&T Coordinator and technical manager on multiple R&T activities for EU, ESA and CNES studies in the fields of data handling and radio-frequency engineering. He also acts as avionics architect and/or radio-frequency architect on multiple satellite advanced phases (A/B1) for Science and Observation missions.

Figure captions:
Fig. 1.  The active diagnosis scheme.
Fig. 2.  Spacewire Network architecture
Fig. 3. Example of a hybrid system.
Fig. 4 ActHyDiag architecture
Fig. 5: AO main Algorithm flowchart
Fig. 6: AO_Algorithm
Fig. 7: Supremica automata for instruments and SpaceWire router
Fig. 8: Original statechart for the Spacewire Network FDIR